\documentclass{article}

\usepackage[preprint]{corl_2026} 

\usepackage{booktabs} 
\usepackage{multirow} 
\usepackage{amsmath}  
\usepackage{graphicx} 
\usepackage{enumitem} 
\usepackage{wrapfig} 
\usepackage{booktabs}
\usepackage[table]{xcolor}
\usepackage{colortbl}
\usepackage{threeparttable}
\usepackage{wrapfig}

\usepackage{pifont}

\definecolor{lineBlue}{RGB}{120,190,225}
\definecolor{tableBlue}{RGB}{232,244,250}

\usepackage{amssymb} 
\usepackage{amsmath}

\usepackage{tabularx} 

\usepackage{bbm}  

\usepackage{adjustbox}
\usepackage{booktabs}
\usepackage{graphicx}
\usepackage[table]{xcolor}

\usepackage{booktabs}
\usepackage{array}
\usepackage{tabularx}
\usepackage[table]{xcolor}

\definecolor{sectiongray}{RGB}{242,242,242}
\definecolor{headgray}{RGB}{225,225,225}
\definecolor{taskblue}{RGB}{220,235,247}
\definecolor{qualitygreen}{RGB}{226,241,226}
\definecolor{diagorange}{RGB}{249,234,214}

\newcolumntype{Y}{>{\centering\arraybackslash}X}

\title{Physics-Guided Biomechanical Gait Adaptation for Humanoid Locomotion on Extreme Sloped Terrains}

\author{
  Xuanyu Chen$^{1,\star}$, Mohan Liu$^{1,\star}$, Dengchen Mei$^1$, Zhihao Gu$^1$, Haitian Zhang$^1$, Kaimin Mao$^1$ \\ \textbf{Haiyue Zhu$^2$, Shijun Yan$^2$, Lin Wang$^{1,\dagger}$}\\
  $^1$Nanyang Technological University \ $^2$A*STAR \ $^\star$Co-first Authors \ $^{\dagger}$Corresponding Author\\
}

\begin{document}
\maketitle


\begin{figure}[h!]
\vspace{-30pt}
    \centering
    \includegraphics[width=.9\linewidth]{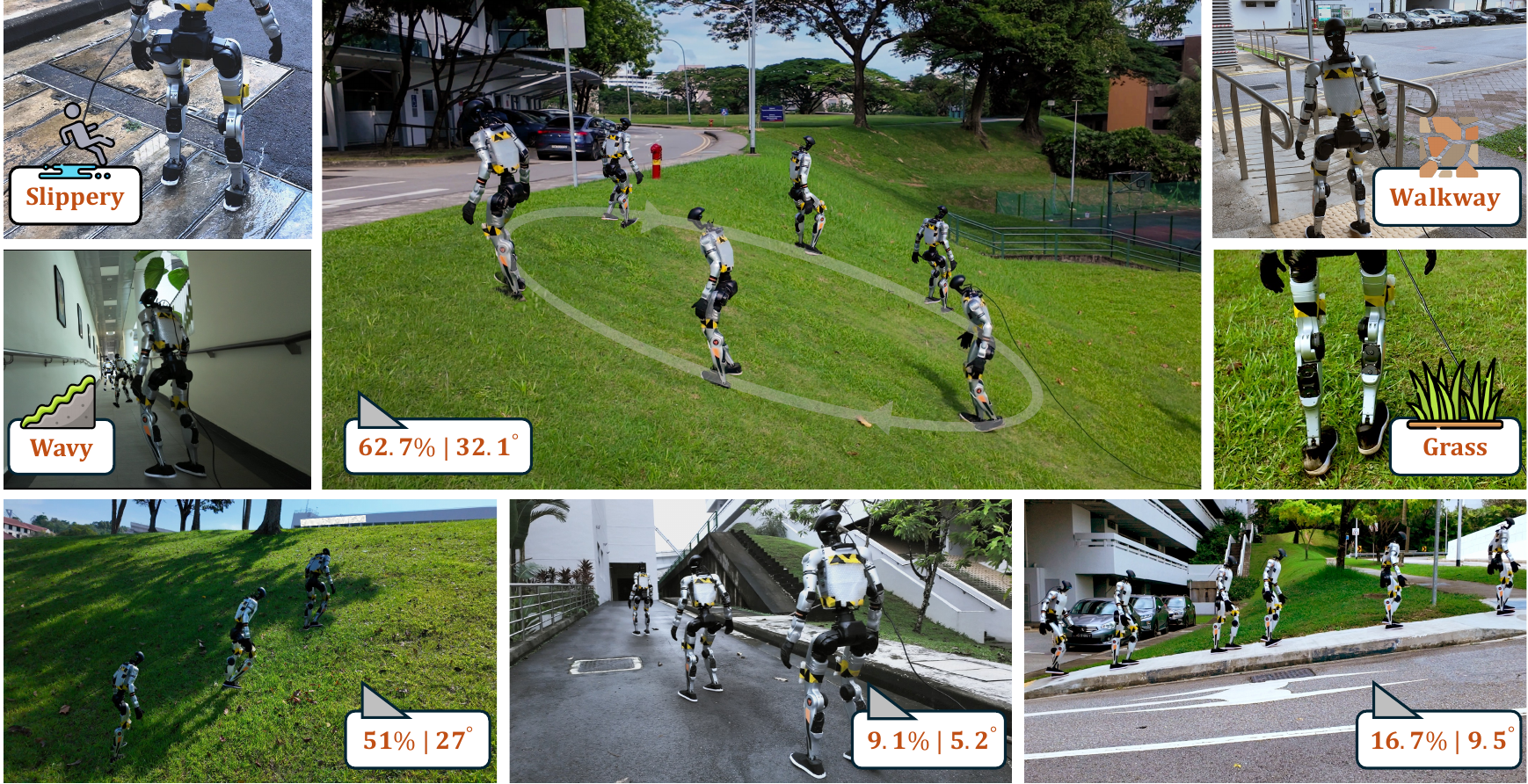}
    \vspace{-5pt}
    \caption{\textbf{Real-world locomotion on sloped terrains}. Our robot traverses grassy slopes up to $62.7\%$ ($32.1^\circ$) grade and generalizes to slippery surfaces, grass, wavy terrains, and level walkways.}
    \label{fig:teaser}
    \vspace{-5pt}
\end{figure}


\begin{abstract}
Model-free reinforcement learning has enabled impressive humanoid locomotion; however, control on steep slopes remains largely unexplored. Unlike flat or discrete terrains, sloped terrains impose a persistent gravitational bias that demands simultaneous \textbf{stability} and \textbf{posture control}. Consequently, under generic reward formulations, policies can converge to slow, conservative low-center-of-mass (CoM) crouched gaits. 
In this work, we propose a novel two-stage physics-guided framework, dubbed \textbf{HumoSlope}, dedicated to robust humanoid locomotion on diverse sloped terrains. Specifically, 
\textbf{Stage I} establishes a terrain-consistent balance prior by introducing a slope-adaptive Zero Moment Point (ZMP) regularizer evaluated directly on the local inclined support plane rather than a world-horizontal reference. To prevent the resulting policy from defaulting to a crouched posture, \textbf{Stage II} introduces the Biomechanical Slope Gait Adapter (\textbf{BSGA}). Utilizing extracted macroscopic terrain descriptors as privileged, training-only signals, BSGA dynamically gates soft reward priors to modulate CoM height and lower-limb coordination based on the estimated slope geometry—encouraging hip-dominant uphill propulsion and knee-oriented downhill braking. Crucially, the deployed actor remains entirely proprioceptive, requiring no online exteroceptive sensing. 
Extensive Sim-to-Real experiments demonstrate that our framework effectively mitigates posture degeneration and enables blind, continuous traversal of outdoor grass slopes up to \textbf{62.7\% ($32.1^\circ$)}, validating a physics-guided approach to challenging slope terrain adaptation.
\end{abstract}

\vspace{-5pt}

\section{Introduction}
In recent years, model-free humanoid locomotion control has undergone a substantial advance thanks to massively parallel simulation and Sim-to-Real reinforcement learning pipelines~\citep{rudin2022learning,hwangbo2019learning,radosavovic2024real}. Recent policies can traverse challenging terrains such as stairs, stepping stones, and parkour-style obstacles~\citep{siekmann2021blind,wang2025beamdojo,zhuang2024humanoid}. However, continuous steep slopes remain underexplored as a dedicated locomotion regime, despite being common in daily environments and outdoor robot deployment, such as ramps, grassy slopes, uneven outdoor paths, and natural hillsides. Prior studies often include inclined surfaces as one case within broader multi-terrain evaluations, however, they provide limited analysis of how slope grade affects traversal limits, posture, joint loading, and gait adaptation~\citep{radosavovic2024learning,gu2024advancing,song2026gait,suliman2025reinforcement}. 

This distinction matters because an incline imposes a critical gravitational bias on the whole body rather than requiring only short-horizon foothold selection or recovery. As the slope increases, the robot must maintain dynamic stability while controlling posture under shifted Center of Mass (CoM) projection, altered support geometry, and sustained lower-limb loading. These effects are further amplified on rough outdoor slopes, where local surface irregularities interact with the global terrain inclination. Hence, the key research question is: \textbf{how can a humanoid maintain simultaneous stability and posture control under a persistent gravitational bias in extreme sloped terrains?}

Under generic reward formulations, \textit{e.g.}, \citep{unitreerl2024}, we observe an undesired low-CoM strategy in model-free humanoid slope locomotion: the policy remains persistently crouched, resembling a ``Groucho gait''~\citep{mcmahon1987groucho,peng2017deeploco}
(see Fig.~\ref{fig:terrain_policy_posture}b), with slower traversal and degraded posture quality.
\textbf{Our key insight} is that this behavior provides only a form of undesired stability. That is, while it may reduce immediate tip-over risk on steep inclines, it causes severe posture degeneration on flat and mild slopes, overuses low CoM stability, increases hip and knee loading, and ultimately caps further slope traversal capabilities. This suggests that robust slope locomotion requires not only\textbf{ dynamic stability}, but also \textbf{slope-conditioned posture control} to effectively mitigate posture degeneration.

To address these challenges, \textbf{our key idea} is to achieve slope-adaptive locomotion balance via asymmetric biomechanical gait modulation. Accordingly, we propose \textbf{HumoSlope}, a novel two-stage physics-guided framework for robust humanoid locomotion on sloped terrains. Specifically, Stage~I introduces a slope-adaptive Zero Moment Point (\textbf{ZMP}) regularizer (Sec. \ref{sec:zmp}). It establishes a baseline terrain-consistent balance prior by evaluating balance deviation on the local inclined support plane rather than a world-horizontal reference~\citep{vukobratovic1972stability,xie2025humanoid}. 
Stage~II then addresses the crouched-gait degeneration by introducing the
Biomechanical Slope Gait Adapter (\textbf{BSGA}) (Sec \ref{sec:bsga}). 
Prior studies in human locomotion~\citep{vernillo2017biomechanics,
roberts2005sources,minetti2002energy} show that uphill and downhill locomotion impose asymmetric mechanical demands: uphill motion requires increased positive work for propulsion with substantial hip involvement, whereas downhill motion involves braking, and negative work, often
with increased knee involvement. 
We translate these biomechanical insights into the novel design cues for the BSGA.  
Our BSGA is a \textbf{training-time adaptation} module conditioned on a five-dimensional macroscopic terrain descriptor obtained by applying principal component analysis (PCA) to the simulator-provided height-scan patch~\cite{pearson1901liii}.
It has two key roles during training. \textbf{First}, it can provide the privileged critic with both macroscopic terrain orientation and local surface details. \textbf{Second}, it allows the biomechanical
insights to be expressed as a compact set of coupled posture-and-gait
priors. In a nutshell, these priors 
encourage hip-dominant
uphill propulsion and knee-oriented downhill braking while keeping the deployed
actor entirely proprioceptive, as demonstrated in Fig.~\ref{fig:teaser}.


In summary, this paper makes three key contributions: (\textbf{I})
We introduce a slope-adaptive ZMP regularizer evaluated on the local inclined support plane to help Stage~I acquire a stable blind locomotion prior with improved balance on slopes. The resulting actor policy warm-starts the subsequent slope-adaptation stage. (\textbf{II}) We propose BSGA that mitigates crouched-gait degeneration while preserving blind deployment by gating core soft priors for CoM height, uphill/downhill gait asymmetry, and swing-leg guidance.  (\textbf{III}) 
We validate our HumoSlope framework on the humanoid in held-out compound slope-track benchmarks and real-world experiments. The learned blind policy traverses simulated compound slopes up to \(36^\circ\) and outdoor grass slopes up to  \(32.1^\circ\), showing superior robustness and slope-conditioned adaptation capacity over baselines (Fig.~\ref{fig:teaser}).







	
\vspace{-10pt}
\section{Related Work}
\vspace{-10pt}
\label{sec:related_work}


\noindent  \textbf{Locomotion on Complex and Sloped Terrains.}
Model-free reinforcement learning has advanced legged and humanoid locomotion across complex terrains, including stairs, discrete stepping stones, and high-agility obstacles~\citep{siekmann2021blind, wang2025beamdojo, zhuang2024humanoid}. These works demonstrate strong terrain traversal by improving contact timing, foothold use, and whole-body coordination. Recent Sim-to-Real methods further show that robust policies can be learned at scale by leveraging proprioceptive feedback, observation histories, and domain randomization~\citep{radosavovic2024real, radosavovic2024learning, xie2020learning, li2025reinforcement}. To manage terrain variations, existing literature is broadly divided into perceptive approaches utilizing exteroceptive sensors~\citep{song2026gait, ben2025gallant, he2025attention, agarwal2023legged, long2025learning}, and blind strategies relying strictly on internal joint states and contact histories~\citep{hwangbo2019learning, sun2025world, kumar2021rma}. However, within these broad multi-terrain benchmarks, slopes are typically included as one test factor among others, rather than a distinct physical challenge~\citep{radosavovic2024learning, gu2024advancing, zhang2026rpl,gu2026learning,gu2026limode}.
\textit{Consequently, slope-specific humanoid locomotion remains highly under-explored.} 
Without slope-specific posture regulation, generic reward formulations may encourage conservative low-CoM crouched strategies that overuse CoM stability.
\textit{Overcoming these challenges requires a physics-guided reformulation of stability combined with adaptive, geometry-dependent biomechanical modulation to maintain traversal-efficient, slope-conditioned posture under proprioceptive deployment.}

\noindent  \textbf{Dynamic Balance and Support Geometry.}
Classical bipedal balance relies on reduced-order criteria -- such as the ZMP and linear inverted pendulum model -- to map CoM trajectories to feasible support regions~\citep{vukobratovic1972stability, sardain2004forces, gu2026humanoid}, often embedded into DRL via reward shaping~\citep{xie2025humanoid}. However, most formulations implicitly assume a flat support plane orthogonal to gravity~\citep{kim2007walking, seven2011humanoid}. On steep inclines, this introduces a severe physical misalignment between the contact plane, support polygon, and gravitational vector. While theoretical multi-contact formulations suggest evaluating balance relative to local contact manifolds~\citep{caron2016zmp, brecelj2022zero}, transforming these geometric insights into robust, end-to-end policy learning remains an open challenge.
\textit{Our HumoSlope addresses this gap by introducing a slope-adaptive ZMP regularizer that evaluates balance deviation on the estimated local support plane.
This balance prior is then combined with Biomechanical Slope Gait Adapter (BSGA), which gates posture and gait reward priors using a macroscopic terrain descriptor.}

\noindent \textbf{Biomechanical Priors for Slope Locomotion.}
Human biomechanics studies show that sloped locomotion is a specialized posture and joint-coordination problem rather than a variant of level-ground walking. Slopes induce direction-dependent changes in trunk posture and joint work, requiring greater positive work for uphill propulsion and increased negative work for downhill braking~\citep{sheehan2012similar, pickle2016functional, dewolf2018kinematic, nuckols2020mechanics, papachatzis2023mechanics}. While learning-based methods introduce gait priors via reference motions, histories, central pattern generators, or world models~\citep{gu2024advancing, kumar2021rma, peng2018deepmimic, yao2022humanoid, fang2025bio, liang2025terrain, jin2025teacher}, they are generally not designed to encode uphill/downhill joint-level redistribution as explicit slope-conditioned priors. 
In our setting, generic survival- and tracking-oriented objectives can encourage persistent low-CoM crouched strategies resembling ``Groucho'' gaits~\citep{mcmahon1987groucho}.
\textit{This motivates BSGA, which uses estimated slope geometry to guide CoM-height
regulation and lower-limb coordination during training.}

\begin{figure}[t!]
    \centering
    \includegraphics[width=.9\textwidth]{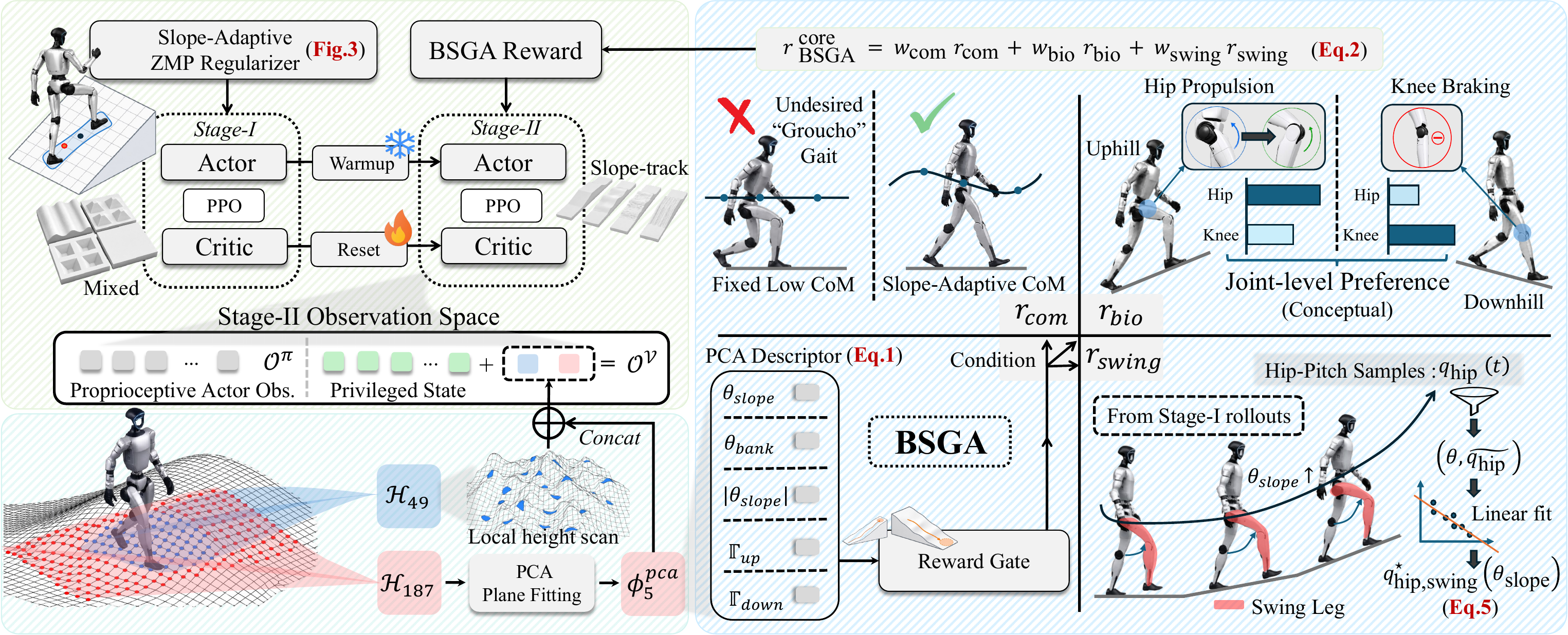}
    \vspace{-5pt}
    \caption{Overview of the proposed two-stage blind slope-locomotion framework. The actor uses only proprioceptive observations at deployment, while the privileged critic and reward gates use terrain descriptor during training. Stage~I learns a slope-adaptive ZMP regularized locomotion prior, and Stage~II adapts it with BSGA for slope-conditioned posture, gait, and swing-leg modulation.}
    \label{fig:main_architecture}
    \vspace{-10pt}
\end{figure}

\vspace{-5pt}
\section{The Proposed HumoSlope Framework}
\vspace{-5pt}
\label{sec:method}
\noindent \textbf{Overview.} Fig.~\ref{fig:main_architecture} provides an overview of our HumoSlope, which is a two-stage framework.
Specifically, Stage~I first trains a blind actor--critic policy on mixed terrains with the slope-adaptive ZMP regularizer, producing a balance-oriented locomotion prior. 
This actor is then used to warm-start Stage~II, while the critic is reset to accommodate the upgraded privileged observation space. The deployed actor remains purely proprioceptive, whereas the training-time critic receives additional privileged state and terrain information, including a compressed height scan and the principal component analysis (PCA) terrain descriptor. 
The Biomechanical Slope Gait Adapter (BSGA) then uses this training-only descriptor to condition the biomechanically motivated soft reward
priors~\cite{vernillo2017biomechanics,minetti2002energy,padulo2013paradigm} for CoM-height regulation, uphill/downhill lower-limb coordination, and swing-leg guidance. 


\textbf{Problem Formulation.} We formulate the locomotion problem as a Partially Observable Markov Decision Process (POMDP)~\cite{KAELBLING199899,ASTROM1965174} with asymmetric actor--critic observations, defined by \(\langle \mathcal{S}, \mathcal{A}, \mathcal{P}, \mathcal{R}, \mathcal{O}^{\pi}, \mathcal{O}^{V}, \gamma \rangle\). 
Here, \(\mathcal{S}\) is the underlying physics state, \(\mathcal{A}\) the action space, \(\mathcal{P}\) the transition dynamics, \(\mathcal{R}\) the reward function, \(\mathcal{O}^{\pi}\) and \(\mathcal{O}^{V}\) the actor and critic observation spaces, and \(\gamma\) the discount factor.
The actor observation \(\mathbf{o}^{\pi}_t\in\mathcal{O}^{\pi}\) contains only deployable
proprioceptive signals, including base angular velocity, projected gravity, velocity commands, joint states, and action history.
Thus, the deployed actor requires no exteroceptive sensing at inference time. During training, the critic observation \(\mathbf{o}^{V}_t\in\mathcal{O}^{V}\) additionally includes privileged state and terrain information, such as the true base linear velocity, a compressed height scan \(\mathcal{H}_{49}\), and the PCA terrain descriptor \(\boldsymbol{\phi}^{\mathrm{PCA}}_5\). 
The action \(\mathbf{a}_t\in\mathcal{A}\) is a joint-position residual target, which is tracked by a low-level PD controller. 
The actor \(\pi_{\theta}\) is optimized with PPO~\citep{schulman2017proximalpolicyoptimizationalgorithms} using a privileged critic \(V_{\phi}\).
We next describe the two training stages.

\subsection{Stage I: Slope-Adaptive ZMP Regularization}
\label{sec:zmp}
\begin{wrapfigure}{r}{0.4\textwidth}
    \vspace{-12pt} 
    \centering
    \includegraphics[width=0.95\linewidth]{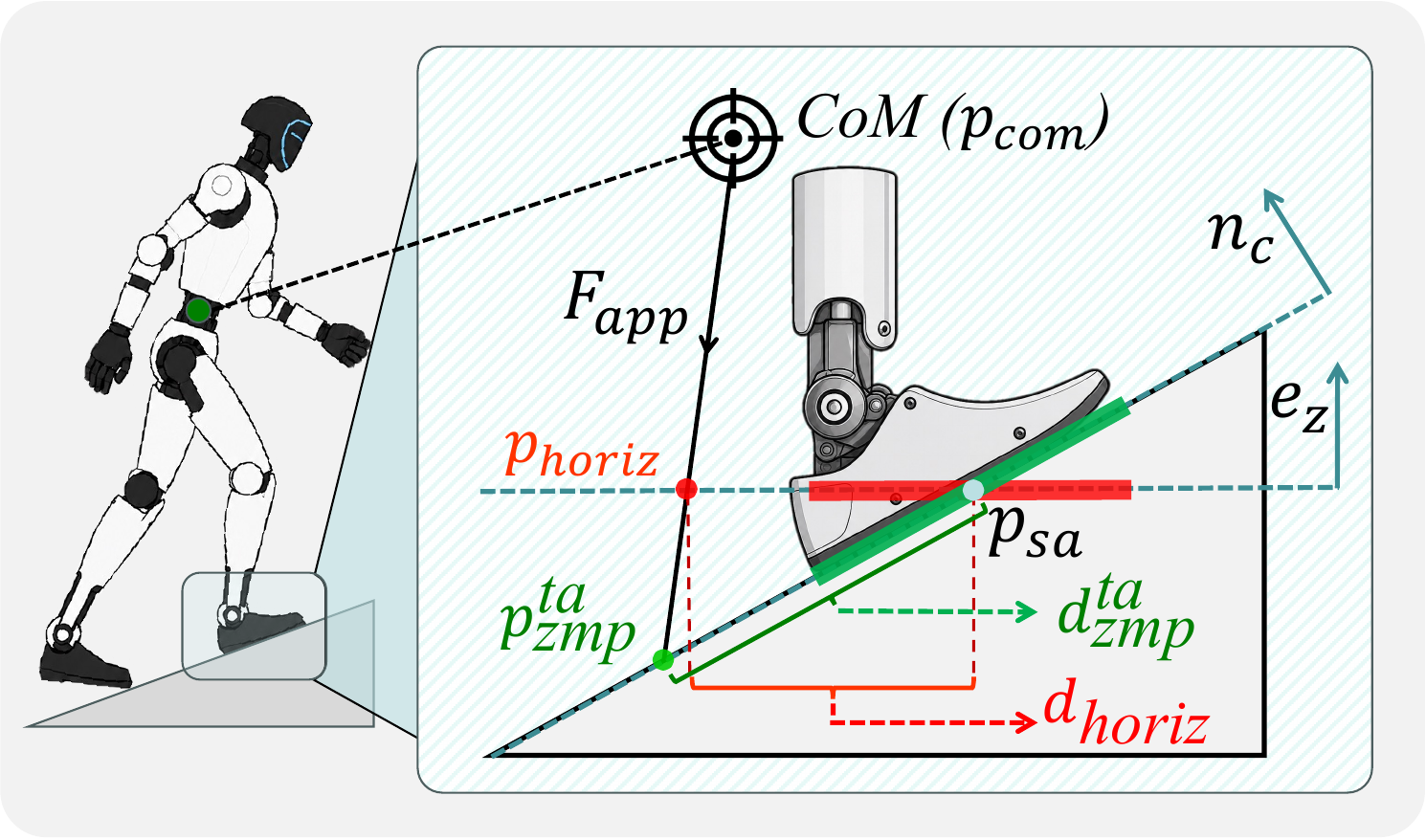}
    \vskip -0.1in
    \caption{On slope, the horizontal-reference deviation \(d_{\mathrm{horiz}}\) can differ from the terrain-aligned deviation \(d_{\mathrm{zmp}}^{\mathrm{ta}}\).}\label{fig:terrain_aligned_zmp}
    \vskip -0.1in
\end{wrapfigure}

Stage~I aims to learn a warm-start locomotion prior. 
Policies trained mainly with generic tracking and survival rewards can traverse moderate slopes, but may develop fragile edge-of-stability gaits on steeper inclines. 
We therefore add a slope-adaptive ZMP regularizer, implemented as a
terrain-aligned ZMP deviation on the estimated local support plane and inspired by classical ZMP stability criteria~\citep{vukobratovic1972stability}, as depicted in Fig.~\ref{fig:terrain_aligned_zmp}.
Existing ZMP/ZML-inspired RL rewards often evaluate balance deviation on a world-horizontal reference plane~\citep{xie2025humanoid}. 
On slopes, however, this reference is misaligned with the actual support surface and can introduce a geometric bias in the measured balance deviation. 
In contrast, our slope-adaptive formulation evaluates this deviation on a local inclined support plane estimated from the stance foot, as illustrated in Fig.~\ref{fig:terrain_aligned_zmp}. 

\noindent \textbf{Approximation scope and support anchor.}
A strict ZMP/ZML reward would require estimating centroidal angular-momentum rates and a reliable contact-wrench support region, which are noisy and contact-solver dependent in massively parallel RL. 
We thus use a point-mass apparent-force surrogate and avoid brittle contact-patch boundaries by defining a contact-force-weighted support anchor
\(\mathbf{p}_{\mathrm{sa}}=((F_L+\varepsilon)\mathbf{p}_L+(F_R+\varepsilon)\mathbf{p}_R)/(F_L+F_R+2\varepsilon)\),
where \(F_L\) and \(F_R\) are the left and right contact-force magnitudes, \(\mathbf{p}_L,\mathbf{p}_R\) are the corresponding foot positions, and \(\varepsilon\) is a small smoothing constant. This load-aware anchor interpolates between the feet according to the measured support forces and serves as a conservative interior support target for the terrain-aligned ZMP deviation.

\noindent \textbf{Terrain-aligned intersection and reward.}
We estimate the local support-plane normal from the dominant stance foot as \(\mathbf{n}_c=\mathbf{R}(\mathbf{q}_{\mathrm{foot}})[0,0,1]^\top\), and use the point-mass apparent specific force \(\mathbf{F}_{\mathrm{app}}=\mathbf{g}-\mathbf{a}_{\mathrm{com}}\), where \(\mathbf{p}_{\mathrm{com}}\) and \(\mathbf{a}_{\mathrm{com}}\) denote the mass-weighted full-body CoM position and acceleration, respectively. 
The terrain-aligned ZMP position is obtained by intersecting the ray
\(\mathbf{p}_{\mathrm{com}}+t\mathbf{F}_{\mathrm{app}}\) with the local support plane \((\mathbf{x}-\mathbf{p}_{\mathrm{sa}})^\top\mathbf{n}_c=0\), where
\(t=((\mathbf{p}_{\mathrm{sa}}-\mathbf{p}_{\mathrm{com}})^\top\mathbf{n}_c)/(\mathbf{F}_{\mathrm{app}}^\top\mathbf{n}_c)\), and
\(\mathbf{p}_{\mathrm{zmp}}^{\mathrm{ta}}=\mathbf{p}_{\mathrm{com}}+t\mathbf{F}_{\mathrm{app}}\).
The deviation from the support anchor is shaped into a dense Stage-I reward
\(r_{\mathrm{zmp}}^{\mathrm{ta}}=\exp(-d_{\mathrm{zmp}}^{\mathrm{ta}}/\sigma_{\mathrm{zmp}})\), where
\(d_{\mathrm{zmp}}^{\mathrm{ta}}=\|\mathbf{p}_{\mathrm{zmp}}^{\mathrm{ta}}-\mathbf{p}_{\mathrm{sa}}\|_2\).
This regularizer is used only during Stage~I to help the actor policy acquire a terrain-consistent balance prior. 

\subsection{Stage II: Biomechanical Slope Gait Adapter (BSGA)}
\label{sec:bsga}
\textbf{Insight.} While Stage~I provides a balance-oriented warm start, slope traversal still requires slope-conditioned posture and gait adaptation. 
The Biomechanical Slope Gait Adapter (BSGA) addresses this issue by using the training-only PCA terrain descriptor to gate a compact set of soft reward priors, as illustrated
in the right part of Fig.~\ref{fig:main_architecture}.

\textbf{PCA terrain descriptor.}
Ray-cast height scans provide local terrain samples that implicitly contain both surface irregularities and the dominant orientation of the scanned patch.
To expose the macroscopic slope structure, we apply PCA to the valid height-scan points and project the resulting plane normal into the robot heading frame.
This yields the longitudinal slope angle $\theta_{\mathrm{slope}}$ and cross-slope tilt angle $\theta_{\mathrm{bank}}$.
We construct the five-dimensional PCA terrain descriptor as:
\begin{equation}
    \boldsymbol{\phi}^{\mathrm{PCA}}_5
    =
    \left(
    \theta_{\mathrm{slope}},
    \theta_{\mathrm{bank}},
    |\theta_{\mathrm{slope}}|,
    \mathbbm{1}_{\mathrm{up}},
    \mathbbm{1}_{\mathrm{down}}
    \right),
    \label{eq:pca_descriptor}
\end{equation}
where $\mathbbm{1}_{\mathrm{up}}$ and $\mathbbm{1}_{\mathrm{down}}$ indicate ascending and descending regimes. Absolute world height is excluded so that the descriptor captures terrain orientation rather than elevation.
In Stage~II, \(\boldsymbol{\phi}^{\mathrm{PCA}}_5\) augments the privileged critic with macroscopic terrain cues and gates the BSGA reward priors, as shown in Fig.~\ref{fig:main_architecture}, while the compressed height scan \(\mathcal{H}_{49}\) preserves local surface details.

\textbf{BSGA core reward.}
Given \(\boldsymbol{\phi}^{\mathrm{PCA}}_5\), BSGA gates a compact set of descriptor-conditioned soft reward priors. 
Rather than imitating human trajectories, these priors encode biomechanically motivated ascent/descent asymmetries in a robot-compatible form. 
The core BSGA reward is
\begin{equation}
    r_{\mathrm{BSGA}}^{\mathrm{core}}
    =
    w_{\mathrm{com}} r_{\mathrm{com}}
    +
    w_{\mathrm{bio}} r_{\mathrm{bio}}
    +
    w_{\mathrm{swing}} r_{\mathrm{swing}} .
    \label{eq:bsga_core}
\end{equation}
Here, $r_{\mathrm{com}}$ regulates slope-conditioned CoM height,
$r_{\mathrm{bio}}$ encodes directional uphill/downhill gait preferences, and
$r_{\mathrm{swing}}$ provides calibration-informed swing-leg guidance. In the full Stage-II objective, they are combined with standard locomotion rewards and auxiliary regularizers for kinematic feasibility and upper-body stability.

The CoM branch discourages persistent low-CoM crouching by replacing a fixed height reference with a slope-conditioned CoM-height target. 
Let \(\rho_{\mathrm{slope}}\in[0,1]\) denote the normalized slope intensity derived from \(|\theta_{\mathrm{slope}}|\). 
The target height is
\begin{equation}
    h_{\mathrm{tgt}}
    =
    h_{\mathrm{nom}}\cos(|\theta_{\mathrm{slope}}|)
    +
    \rho_{\mathrm{slope}}
    \left(
        b_{\mathrm{up}}\mathbbm{1}_{\mathrm{up}}
        +
        b_{\mathrm{down}}\mathbbm{1}_{\mathrm{down}}
    \right),
    \label{eq:com_height}
\end{equation}
where \(h_{\mathrm{nom}}\) is the flat-terrain CoM-height reference, and \(b_{\mathrm{up}}\), \(b_{\mathrm{down}}\) are asymmetric offsets for ascent and descent. 
The reward \(r_{\mathrm{com}}\) softly penalizes the CoM-height deviation from \(h_{\mathrm{tgt}}\).

The biomechanical branch encodes ascent/descent asymmetry as a low-weight directional gait prior, motivated by human slope-locomotion studies~\citep{vernillo2017biomechanics,minetti2002energy}:
\begin{equation}
    r_{\mathrm{bio}}
    =
    \frac{
        w_{\mathrm{up}} r_{\mathrm{hip}}
        +
        w_{\mathrm{down}} r_{\mathrm{down}}
    }{
        w_{\mathrm{up}}+w_{\mathrm{down}}
    },
    \qquad
    r_{\mathrm{down}}
    =
    \lambda_{\mathrm{brake}} r_{\mathrm{brake}}
    +
    \lambda_{\mathrm{stride}} r_{\mathrm{stride}} .
    \label{eq:bio_prior}
\end{equation}

Here, \(r_{\mathrm{hip}}\) biases ascent toward hip-dominant propulsion, while \(r_{\mathrm{down}}\) combines controlled knee-oriented braking and reduced over-stepping for descent.
The ascent/descent terms in \(r_{\mathrm{bio}}\) are gated by \(\mathbbm{1}_{\mathrm{up}}\) and \(\mathbbm{1}_{\mathrm{down}}\).

The swing branch provides a calibrated slope-dependent hip-pitch reference during training:
\begin{equation}
    q_{\mathrm{hip,swing}}^\star(\theta_{\mathrm{slope}})
    =
    \beta_0
    +
    \beta_1
    \mathrm{clip}
    \left(
        \theta_{\mathrm{slope}},
        0,
        \theta_{\mathrm{clip}}
    \right).
    \label{eq:swing_calibration}
\end{equation}
The coefficients are fitted from Stage-I rollouts using a contact-force-filtered upper-percentile hip-pitch trend across slope angles. 
The reward \(r_{\mathrm{swing}}\) softly tracks this target during swing without imposing a hard trajectory.

\noindent \textbf{Two-stage training summary.} As summarized on the left side of Fig.~\ref{fig:main_architecture}, Stage~I learns a balance-oriented prior on mixed terrains, whereas Stage~II switches to slope-track terrains and uses BSGA for posture and gait adaptation. 
The Stage-I actor is reused to preserve locomotion capability, while the critic is reset to match the new terrain distribution, privileged observations, and reward objective.

\section{Experiments}
\label{sec:result}

\begin{figure}[t]
    \centering
    \includegraphics[width=.9\textwidth]{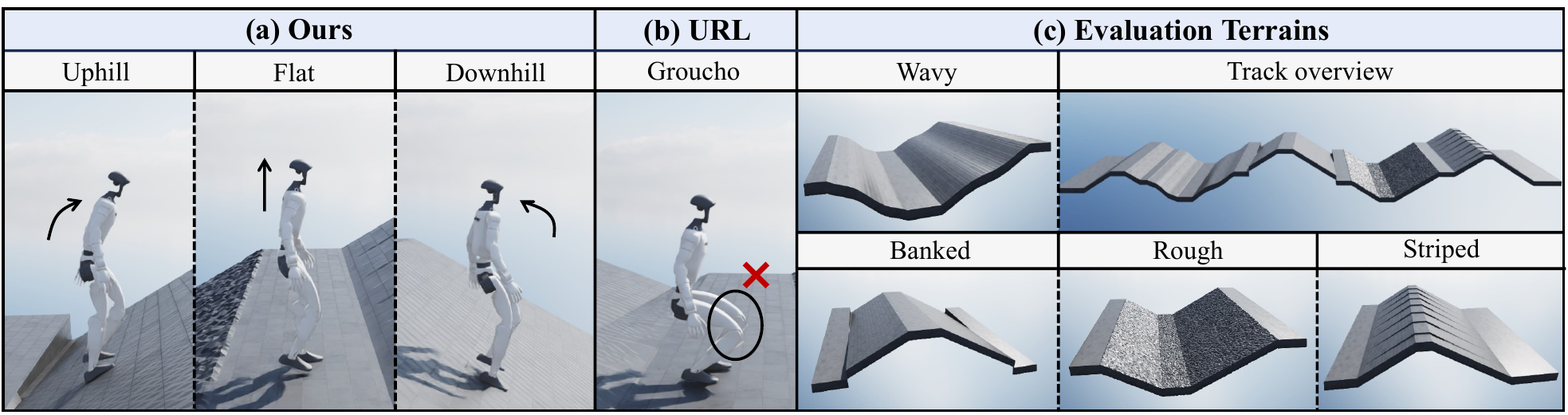}
    \vspace{-5pt}
    \caption{Held-out compound slope-track benchmark and representative policy behaviors. (a) Our policy exhibits different whole-body postures across terrain segments: forward torso lean during ascent, near-upright posture on flat terrain, and slight backward torso lean during descent. (b) The Unitree RL Lab (URL) baseline exhibits a crouched Groucho gait. (c) The benchmark combines compound slope segments with wavy, banked, rough, and striped terrain variants.}\label{fig:terrain_policy_posture}
    \vskip -0.1in
\end{figure}
\vspace{-5pt}
\subsection{Experimental Setup}
\label{subsec:setup}
\vspace{-5pt}
\textbf{Implementation and baselines.}
All simulation experiments are conducted in Isaac Lab~\citep{mittal2025isaaclab} on the Unitree~G1 humanoid. 
HumoSlope is trained with PPO using thousands of parallel environments on a single NVIDIA RTX~5090 GPU, with domain randomization for sim-to-real robustness. 
Stage~I and Stage~II are trained until validation performance saturates. 
We compare against Unitree RL Lab (URL)~\citep{unitreerl2024}, FastTD3~\citep{seo2025fasttd3}, and Gallant~\citep{ben2025gallant}. 
URL and FastTD3 are proprioception-only baselines, while Gallant uses exteroceptive terrain sensing. 
For fairness, all baselines are retrained with their original reward functions and policy architectures, and their training terrain distributions are extended, where compatible, to closely match our slope range and terrain types.
For Table~\ref{tab:main_results}, each method is evaluated using three post-saturation checkpoints, and results are averaged across checkpoints and friction tiers.

\textbf{Held-out slope-track benchmark.}
We evaluate each method on a 35\,m held-out compound slope track. For a nominal grade \(\theta\), the track consists of an irregular sequence of flat, uphill, downhill, and plateau segments; inclined segments use slopes of \(\pm\theta\).
The surface type also changes along the track among five variants: smooth, wavy, banked, rough, and striped slopes, as visualized in Fig.~\ref{fig:terrain_policy_posture}. This benchmark is intentionally different from the Stage~II training terrain to test generalization rather than memorization of a fixed track. We evaluate four nominal grades \(\theta\):
0\%~($0^\circ$), 17.6\%~($10^\circ$), 36.4\%~($20^\circ$), and
57.7\%~($30^\circ$). For each nominal grade, we test three friction tiers defined as \(\mu=\tan(|\theta|)+\Delta\), where \(\Delta\in\{0.3,0.6,0.9\}\). This normalizes the friction margin relative to the static sliding threshold of each slope, making the low/mid/high tiers physically comparable across grades. A trial is successful if the robot reaches 31.5\,m, corresponding to 90\% of the track length. The timeout is set to 140\,s, twice the ideal traversal time under the 0.5\,m/s command.

\begin{table}[t]
\centering
\scriptsize
\setlength{\tabcolsep}{2.2pt}
\setlength{\abovecaptionskip}{2pt}
\setlength{\belowcaptionskip}{2pt}
\caption{
Performance across compound slope-track benchmarks.
Each grade is reported as slope ratio with the corresponding angle in parentheses. 
SR is reported in percent, MXD in meters, and \(T_{\mathrm{trav}}\) in seconds. Results are averaged over three checkpoints and three friction tiers.
\textsuperscript{\(\dagger\)}Gallant uses exteroceptive depth sensing; all other
methods are proprioception-only.
}
\label{tab:main_results}
\resizebox{\textwidth}{!}{%
{
\arrayrulecolor{lineBlue}
\begin{tabular}{l l ccc ccc ccc ccc c}
\toprule[0.9pt]
\rowcolor{tableBlue}
\textbf{Method} & \textbf{Input}
& \multicolumn{3}{c}{\textbf{0\% ($0^\circ$)}}
& \multicolumn{3}{c}{\textbf{17.6\% ($10^\circ$)}}
& \multicolumn{3}{c}{\textbf{36.4\% ($20^\circ$)}}
& \multicolumn{3}{c}{\textbf{57.7\% ($30^\circ$)}}
& \textbf{Max Grade}$\uparrow$ \\
\cmidrule(lr){3-5}
\cmidrule(lr){6-8}
\cmidrule(lr){9-11}
\cmidrule(lr){12-14}
\rowcolor{tableBlue}
&
& SR$\uparrow$ & MXD$\uparrow$ & $T_{\mathrm{trav}}\downarrow$
& SR$\uparrow$ & MXD$\uparrow$ & $T_{\mathrm{trav}}\downarrow$
& SR$\uparrow$ & MXD$\uparrow$ & $T_{\mathrm{trav}}\downarrow$
& SR$\uparrow$ & MXD$\uparrow$ & $T_{\mathrm{trav}}\downarrow$
& \\
\midrule[0.4pt]
URL~\citep{unitreerl2024}
& Prop.
& 99.7 & 32.13 & 88.0
& 97.7 & 31.66 & 92.7
& 37.5 & 17.94 & 99.4
& 0.0  & 2.18  & --
& \(47\% (25^\circ)\) \\

FastTD3~\citep{seo2025fasttd3}
& Prop.
& 100.0 & 31.50 & 76.1
& 100.0 & 31.49 & 73.2
& 77.3  & 28.32 & 87.5
& 0.0   & 2.02  & --
& \(70\% (35^\circ)\) \\

Gallant\textsuperscript{\(\dagger\)}~\citep{ben2025gallant}
& Depth
& 99.7 & 31.47 & 51.9
& 90.7 & 30.29 & 53.5
& 78.9 & 28.37 & 54.6
& 0.0  & 6.11  & --
& \(60\%(31^\circ)\) \\

\midrule[0.4pt]
\textbf{Ours}
& Prop.
& \textbf{100.0} & \textbf{32.19} & \textbf{44.3}
& \textbf{100.0} & \textbf{32.19} & \textbf{47.0}
& \textbf{100.0} & \textbf{32.18} & \textbf{48.6}
& \textbf{77.1}  & \textbf{27.54} & \textbf{62.8}
& $\mathbf{73\% (36^\circ)}$ \\
\bottomrule[0.9pt]
\end{tabular}
\arrayrulecolor{black}
}%
}
\vspace{-1.5em}
\end{table}

\textbf{Metrics.}
We report success rate (SR), mean maximum forward distance (MXD), and successful-trial traversal time \(T_{\mathrm{trav}}\) as the main benchmark metrics.
MXD is averaged over all trials, while \(T_{\mathrm{trav}}\) is averaged over successful trials using success-count weighting across conditions.
The Max Grade column reports an additional slope-limit sweep and is not averaged with the four nominal benchmark grades.
For ablations, we also track cost of transport (CoT), yaw-rate variance \(\sigma_{\mathrm{yaw}}^2\), mean CoM height \(\bar{h}_{\mathrm{com}}\), and peak knee torque \(\tau_{\mathrm{knee}}^{\mathrm{pk}}\).
These additional metrics are interpreted jointly with task success as efficiency, stability, posture, and load diagnostics rather than standalone objectives.

\vspace{-5pt}
\subsection{Main Results}\label{subsec:comparison}
\vspace{-5pt}
\textbf{Simulation Results.} 
Table~\ref{tab:main_results} compares our method with both proprioceptive and exteroceptive baselines on the held-out compound slope-track benchmark.
SR measures full-track completion, while MXD captures partial progress before failure. 
Across \(0^\circ\)--\(20^\circ\), our method achieves near-perfect success and the shortest successful-trial traversal time among all baselines.
At the most challenging 57.7\%~($30^\circ$) grade, all baselines fail to complete the track, whereas our policy still achieves \(77.1\%\) SR and reaches an average distance of 27.54\,m.
In the additional slope-limit sweep, our method also reaches the highest maximum grade, \(73\%\)~($36^\circ$).
These results indicate that the proposed two-stage training strategy and descriptor-gated slope adaptation improve robustness on compound uphill/downhill tracks, rather than overfitting to a single fixed ramp.
Qualitatively, although the URL baseline can traverse mild slopes, it tends to adopt a conservative low-CoM posture on flat and mildly inclined segments, consistent with the undesired low-CoM strategy discussed above.
In contrast, our policy maintains a more upright posture on flat terrain and shows slope-varying torso and leg behavior on inclined segments, as shown in Fig.~\ref{fig:terrain_policy_posture}.

\begin{figure}[t]
    \centering
    \includegraphics[width=.9\linewidth]{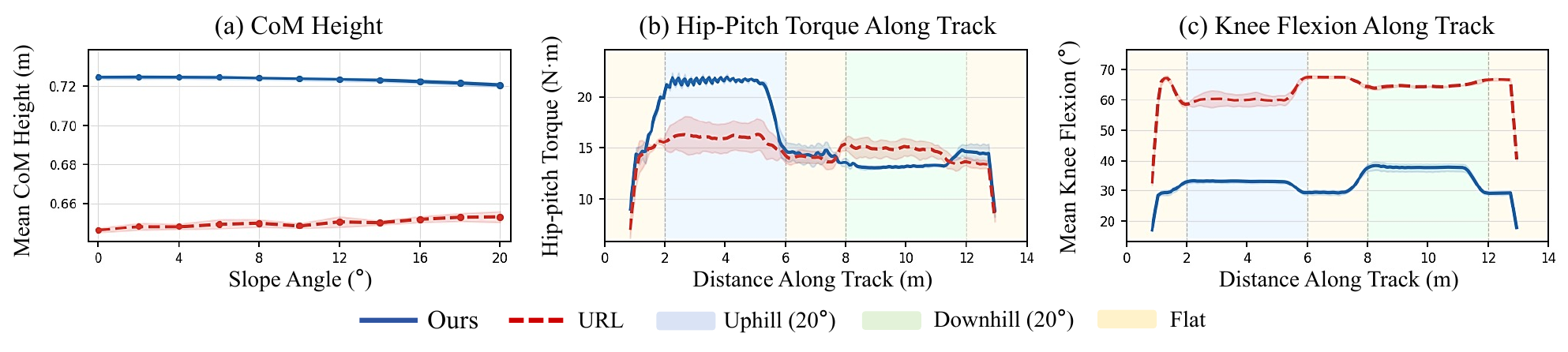}
    \vspace{-5pt}
    \caption{Biomechanical diagnostics. HumoSlope maintains a higher CoM height with mild slope-dependent modulation, while URL remains in a lower-CoM crouched posture. On the \(20^\circ\) track, HumoSlope shows higher uphill hip-pitch torque and increases knee flexion during descent while remaining less crouched than URL, consistent with slope-conditioned lower-limb coordination.
}
    \label{fig:biomech_diagnostics}
    \vskip -0.1in
\end{figure}

Fig.~\ref{fig:biomech_diagnostics} further quantifies the posture and lower-limb trends behind these qualitative behaviors. 
The diagnostics support the same pattern: HumoSlope avoids the persistent low-CoM crouch observed in URL while showing slope-conditioned lower-limb modulation.

\textbf{Real-world experiments.}
We deploy the final HumoSlope policy on the Unitree~G1 humanoid across six outdoor terrains: an asphalt slope, a wavy slope, a roadside slope, two grass slopes, and a daily-use flat walkway.
The deployed actor uses the same proprioceptive observations as in simulation, without online exteroceptive terrain sensing.
The measured mean slope angles range from \(5.2^\circ\) on the asphalt slope to \(32.1^\circ\) on the steeper grass slope, with local grass-slope measurements reaching \(36.4^\circ\). As shown in Fig.~\ref{fig:teaser}, HumoSlope achieves blind continuous traversal on outdoor grass slopes and also traverses asphalt and wavy slopes under wet conditions after rainfall.


\begin{wrapfigure}[5]{r}{0.48\textwidth}
    \vspace{-14pt}
    \centering
    \includegraphics[width=0.98\linewidth]{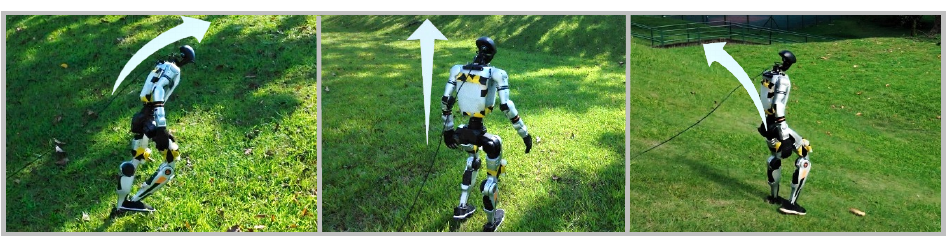}
    \vspace{-10pt}
    \caption{
    Real-world posture on grass terrain.
    }
    \label{fig:realworld_posture}
\end{wrapfigure}

Fig.~\ref{fig:realworld_posture} further shows that the same policy exhibits different postures across grass-terrain segments, with forward lean during ascent, near-upright walking on flat ground, and slight backward lean during descent.
These results suggest that the training-time descriptor-gated adaptation learned in simulation transfers to real outdoor slopes while preserving fully proprioceptive deployment.

\vspace{-5pt}
\subsection{Ablation Study}
\label{subsec:ablation}
Since the individual BSGA reward terms are designed as coupled posture-and-gait priors, we ablate them jointly as descriptor-conditioned reward priors, and separately test the PCA critic cue.
A separately trained ablation suite is evaluated on the held-out compound slope-track benchmark at \(36.4\%\)~($20^\circ$), with its Full model used as the reference for all ablated variants. 
We choose this grade because it exposes differences in traversal robustness, crouched-posture tendency, and lower-limb loading, while remaining informative for most ablated policies. The evaluation uses the same compound segment layout, surface variants, and friction-tier averaging as the main benchmark.

\textbf{Ablation Results and Analysis.} Results in Table~\ref{tab:ablation_study} show that Stage~I alone already provides a strong balance-oriented locomotion prior, reaching \(100.0\%\) SR at \(20^\circ\). 
However, it does so with a much lower CoM height and a longer traversal time, indicating a conservative crouched strategy rather than fast, upright slope-conditioned posture adaptation. 
The full model converts this warm-start prior into a faster and more upright gait. 
Removing the slope-adaptive ZMP regularizer substantially reduces SR and MXD, showing that the Stage-I balance prior remains important for later adaptation. 
In Stage~II, removing either the BSGA critic cue or reward priors degrades robustness, traversal speed, or directional stability, while removing the whole BSGA causes complete failure. 
The posture/load diagnostics further show that the failed w/o BSGA policy has low CoM height and the largest peak knee torque, consistent with low-CoM posture degeneration and uncontrolled lower-limb loading.
\vspace{-5pt}

\begin{table}[t]
\centering

\caption{
Ablation metrics. SR, MXD, and \(T_{\mathrm{trav}}\) follow the definitions in Table~\ref{tab:main_results}.
Additional columns report diagnostics: CoT and \(\sigma_{\mathrm{yaw}}^2\) for efficiency/stability, and \(\bar{h}_{\mathrm{com}}\) and \(\tau_{\mathrm{knee}}^{\mathrm{pk}}\) for posture/load.
Diagnostic metrics are interpreted jointly with task success. 
Stage~I only is evaluated directly after Stage~I, without Stage-II training, whereas w/o BSGA is still trained in Stage~II after removing the BSGA adaptation components (both the BSGA critic cue and BSGA reward priors).
}
\label{tab:ablation_study}
\scriptsize
\setlength{\tabcolsep}{3.8pt}
\renewcommand{\arraystretch}{0.88}

{
\arrayrulecolor{lineBlue}
\begin{tabular*}{.9\textwidth}{
@{\extracolsep{\fill}}
l
ccc
cc
cc
@{}
}
\toprule[0.9pt]
{\rlap{\color{tableBlue}\rule[-0.35em]{0.9\textwidth}{1.45em}}\textbf{Variant}}
&
\multicolumn{3}{c}{\textbf{Task Performance}}
&
\multicolumn{2}{c}{\textbf{Efficiency / Stability}}
&
\multicolumn{2}{c}{\textbf{Posture / Load}}
\\
\cmidrule(lr){2-4}
\cmidrule(lr){5-6}
\cmidrule(l){7-8}
{\rlap{\color{tableBlue}\rule[-0.35em]{0.9\textwidth}{1.45em}}}
&
SR$\uparrow$
&
MXD$\uparrow$
&
$T_{\mathrm{trav}}\downarrow$
&
CoT$\downarrow$
&
$\sigma_{\mathrm{yaw}}^2\downarrow$
&
$\bar{h}_{\mathrm{com}}$
&
$\tau_{\mathrm{knee}}^{\mathrm{pk}}$
\\
\midrule[0.4pt]

\textbf{Full}
& 98.2 & 31.75 & 53.7
& 1.380 & 0.1008
& 0.669 & 117.9 \\

Stage~I only (no Stage~II)
& 100.0 & 32.20 & 67.6
& 0.925 & 0.0148
& 0.535 & 82.9 \\

\midrule[0.4pt]

w/o slope-adaptive ZMP
& 55.6 & 22.08 & 54.1
& 1.906 & 0.3048
& 0.720 & 115.5 \\

\midrule[0.4pt]

w/o BSGA critic cue
& 93.2 & 30.56 & 74.1
& 1.676 & 0.2009
& 0.727 & 83.4 \\

w/o BSGA reward priors
& 26.9 & 18.09 & 71.9
& 1.603 & 0.2031
& 0.722 & 108.9 \\

w/o BSGA
& 0.0 & 9.59 & --
& 1.621 & 0.3133
& 0.659 & 136.4 \\

\bottomrule[0.9pt]
\end{tabular*}
\arrayrulecolor{black}
}
\end{table}
\section{Conclusion, Limitations and Future Work}
\vspace{-5pt}
\label{sec:conclusion}
\textbf{Conclusion.}
We presented \textbf{HumoSlope}, a two-stage physics-guided framework for blind humanoid locomotion on continuous steep slopes.
Stage~I uses a slope-adaptive ZMP regularizer to learn a terrain-consistent balance prior, while Stage~II introduces BSGA to gate biomechanically motivated posture, gait, and swing-leg priors using a training-only PCA terrain descriptor.
The deployed actor remains purely proprioceptive. 
Experiments on the Unitree~G1 show improved robustness and posture adaptation in simulation, as well as blind real-world traversal on outdoor grass slopes up to \(32.1^\circ\). 
These results suggest that combining terrain-consistent physical priors with descriptor-gated biomechanical reward adaptation is an effective route toward robust humanoid slope locomotion.

\textbf{Limitations and Future Work.}
The deployed actor is blind and cannot anticipate upcoming slope transitions,
so posture adjustment only emerges after proprioceptive interaction with the
terrain. This may limit performance on abrupt obstacles, deformable surfaces, or
highly irregular outdoor terrain. Future work will explore adding optional
visual perception for look-ahead terrain cues while preserving HumoSlope's
proprioceptive slope-adaptation capability.



\clearpage
\bibliography{example}  

\clearpage 
\appendix  

\numberwithin{equation}{section}
\numberwithin{figure}{section}
\numberwithin{table}{section}

\end{document}